%% file: main_initialSubmission.tex
\documentclass[10pt,letterpaper]{article}

\usepackage{iccv}
\usepackage{times}
\usepackage{epsfig}
\usepackage{graphicx}
\usepackage{amsmath}
\usepackage{amssymb}
\usepackage{booktabs}
\usepackage[numbers,sort,compress]{natbib}

\usepackage[pagebackref=true,breaklinks=true,letterpaper=true,colorlinks,bookmarks=false]{hyperref}

\iccvfinalcopy 

\begin{document}



\title{Crowd Transformer Network}

\author{Viresh Ranjan\\
Dept. of Computer Science\\
Stony Brook University\\
\and
Mubarak Shah\\
Dept. of Computer Science\\
University of Central Florida\\
\and
Minh Hoai Nguyen\\
Dept. of Computer Science\\
Stony Brook University\\
}

\input{definitions}

\maketitle


\begin{abstract}
In this paper, we tackle the problem of Crowd Counting, and present a crowd density estimation based approach for obtaining the crowd count. Most of the existing crowd counting approaches rely on local features for estimating 
the crowd density map. In this work, we investigate the usefulness of combining local with non-local features for crowd counting. We use convolution layers for extracting local features, and a type of self-attention mechanism for extracting non-local features. We combine the local and the non-local features, and use it for estimating crowd density map. We conduct experiments on three publicly available Crowd Counting datasets, and achieve significant improvement over the previous approaches.
\end{abstract}

\input{Introduction}

\input{Related}

\input{Proposed}
\input{Results}

\section{Conclusion}
We have presented an end-to-end approach for Crowd Counting which combines local and non-local features for estimating crowd density maps. We made use of self-attention mechanism to compute the non-local features. We proposed a novel self-attention approach, which helped in obtaining more accurate crowd density maps. We showed the usefulness of combining local with non-local information for Crowd Counting by surpassing the performance of previous state-of-the-art approaches.

{\small
\setlength{\bibsep}{0pt}
\bibliographystyle{abbrvnat}
\bibliography{main_initialSubmission}
}

\end{document}

%% file: definitions.tex
\def\mA{\mathcal{A}}
\def\mB{\mathcal{B}}
\def\mC{\mathcal{C}}
\def\mD{\mathcal{D}}
\def\mE{\mathcal{E}}
\def\mF{\mathcal{F}}
\def\mG{\mathcal{G}}
\def\mH{\mathcal{H}}
\def\mI{\mathcal{I}}
\def\mJ{\mathcal{J}}
\def\mK{\mathcal{K}}
\def\mL{\mathcal{L}}
\def\mM{\mathcal{M}}
\def\mN{\mathcal{N}}
\def\mO{\mathcal{O}}
\def\mP{\mathcal{P}}
\def\mQ{\mathcal{Q}}
\def\mR{\mathcal{R}}
\def\mS{\mathcal{S}}
\def\mT{\mathcal{T}}
\def\mU{\mathcal{U}}
\def\mV{\mathcal{V}}
\def\mW{\mathcal{W}}
\def\mX{\mathcal{X}}
\def\mY{\mathcal{Y}}
\def\mZ{\mathcal{Z}}

\def\1n{\mathbf{1}_n}
\def\0{\mathbf{0}}
\def\1{\mathbf{1}}

\def\A{{\bf A}}
\def\B{{\bf B}}
\def\C{{\bf C}}
\def\D{{\bf D}}
\def\E{{\bf E}}
\def\F{{\bf F}}
\def\G{{\bf G}}
\def\H{{\bf H}}
\def\I{{\bf I}}
\def\J{{\bf J}}
\def\K{{\bf K}}
\def\L{{\bf L}}
\def\M{{\bf M}}
\def\N{{\bf N}}
\def\O{{\bf O}}
\def\P{{\bf P}}
\def\Q{{\bf Q}}
\def\R{{\bf R}}
\def\S{{\bf S}}
\def\T{{\bf T}}
\def\U{{\bf U}}
\def\V{{\bf V}}
\def\W{{\bf W}}
\def\X{{\bf X}}
\def\Y{{\bf Y}}
\def\Z{{\bf Z}}

\def\a{{\bf a}}
\def\b{{\bf b}}
\def\c{{\bf c}}
\def\d{{\bf d}}
\def\e{{\bf e}}
\def\f{{\bf f}}
\def\g{{\bf g}}
\def\h{{\bf h}}
\def\i{{\bf i}}
\def\j{{\bf j}}
\def\k{{\bf k}}
\def\l{{\bf l}}
\def\m{{\bf m}}
\def\n{{\bf n}}
\def\o{{\bf o}}
\def\p{{\bf p}}
\def\q{{\bf q}}
\def\r{{\bf r}}
\def\s{{\bf s}}
\def\t{{\bf t}}
\def\u{{\bf u}}
\def\v{{\bf v}}
\def\w{{\bf w}}
\def\x{{\bf x}}
\def\y{{\bf y}}
\def\z{{\bf z}}

\def\balpha{\mbox{\boldmath{$\alpha$}}}
\def\bbeta{\mbox{\boldmath{$\beta$}}}
\def\bdelta{\mbox{\boldmath{$\delta$}}}
\def\bgamma{\mbox{\boldmath{$\gamma$}}}
\def\blambda{\mbox{\boldmath{$\lambda$}}}
\def\bsigma{\mbox{\boldmath{$\sigma$}}}
\def\btheta{\mbox{\boldmath{$\theta$}}}
\def\bomega{\mbox{\boldmath{$\omega$}}}
\def\bxi{\mbox{\boldmath{$\xi$}}}
\def\bnu{\mbox{\boldmath{$\nu$}}}                                  
\def\bphi{\mbox{\boldmath{$\phi$}}}
\def\bmu{\mbox{\boldmath{$\mu$}}}

\def\bDelta{\mbox{\boldmath{$\Delta$}}}
\def\bOmega{\mbox{\boldmath{$\Omega$}}}
\def\bPhi{\mbox{\boldmath{$\Phi$}}}
\def\bLambda{\mbox{\boldmath{$\Lambda$}}}
\def\bSigma{\mbox{\boldmath{$\Sigma$}}}
\def\bGamma{\mbox{\boldmath{$\Gamma$}}}

\newcommand{\myminimum}[1]{\mathop{\textrm{minimum}}_{#1}}
\newcommand{\mymaximum}[1]{\mathop{\textrm{maximum}}_{#1}}    
\newcommand{\mymin}[1]{\mathop{\textrm{minimize}}_{#1}}
\newcommand{\mymax}[1]{\mathop{\textrm{maximize}}_{#1}}
\newcommand{\mymins}[1]{\mathop{\textrm{min.}}_{#1}}
\newcommand{\mymaxs}[1]{\mathop{\textrm{max.}}_{#1}}  
\newcommand{\myargmin}[1]{\mathop{\textrm{argmin}}_{#1}} 
\newcommand{\myargmax}[1]{\mathop{\textrm{argmax}}_{#1}} 
\newcommand{\myst}{\textrm{s.t. }}

\newcommand{\denselist}{\itemsep -1pt}
\newcommand{\sparselist}{\itemsep 1pt}

\definecolor{pink}{rgb}{0.9,0.5,0.5}
\definecolor{purple}{rgb}{0.5, 0.4, 0.8}   
\definecolor{gray}{rgb}{0.3, 0.3, 0.3}
\definecolor{mygreen}{rgb}{0.2, 0.6, 0.2}

\newcommand{\cyan}[1]{\textcolor{cyan}{#1}}
\newcommand{\red}[1]{\textcolor{red}{#1}}  
\newcommand{\blue}[1]{\textcolor{blue}{#1}}
\newcommand{\magenta}[1]{\textcolor{magenta}{#1}}
\newcommand{\pink}[1]{\textcolor{pink}{#1}}
\newcommand{\green}[1]{\textcolor{green}{#1}} 
\newcommand{\gray}[1]{\textcolor{gray}{#1}}    
\newcommand{\mygreen}[1]{\textcolor{mygreen}{#1}}    
\newcommand{\purple}[1]{\textcolor{purple}{#1}}       

\definecolor{greena}{rgb}{0.4, 0.5, 0.1}
\newcommand{\greena}[1]{\textcolor{greena}{#1}}

\definecolor{bluea}{rgb}{0, 0.4, 0.6}
\newcommand{\bluea}[1]{\textcolor{bluea}{#1}}
\definecolor{reda}{rgb}{0.6, 0.2, 0.1}
\newcommand{\reda}[1]{\textcolor{reda}{#1}}

\def\changemargin#1#2{\list{}{\rightmargin#2\leftmargin#1}\item[]}
\let\endchangemargin=\endlist
                                               
\newcommand{\cm}[1]{}

\newcommand{\mtodo}[1]{{\color{red}$\blacksquare$\textbf{[TODO: #1]}}}
\newcommand{\myheading}[1]{\vspace{1ex}\noindent \textbf{#1}}
\newcommand{\htimesw}[2]{\mbox{$#1$$\times$$#2$}}


\newif\ifshowsolution
\showsolutiontrue

\ifshowsolution  
\newcommand{\Comment}[1]{\paragraph{\bf $\bigstar $ COMMENT:} {\sf #1} \bigskip}
\newcommand{\Solution}[2]{\paragraph{\bf $\bigstar $ SOLUTION:} {\sf #2} }
\newcommand{\Mistake}[2]{\paragraph{\bf $\blacksquare$ COMMON MISTAKE #1:} {\sf #2} \bigskip}
\else
\newcommand{\Solution}[2]{\vspace{#1}}
\fi

\newcommand{\truefalse}{
\begin{enumerate}
	\item True
	\item False
\end{enumerate}
}

\newcommand{\yesno}{
\begin{enumerate}
	\item Yes
	\item No
\end{enumerate}
}

%% file: Introduction.tex
\section{Introduction}
\begin{figure}[!t]
\centering
\includegraphics[width=\linewidth]{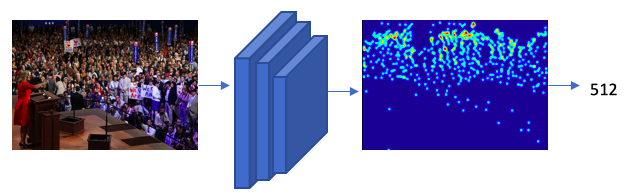}
	\caption{
     \textbf{CNN based crowd density estimation:} Most recent CNN based approaches used a fully convolutional CNN architecture to map a crowd image to the density map, and the crowd count is obtained by summing across all the pixels in the predicted map. These approaches take into account the local information while estimating the crowd density map.
     \label{fig:DensityCNN}}	
\end{figure}

Given an image of a crowd, Crowd Counting, as the name suggests, aims at estimating the total number of people in the image. Crowd Counting has applications in various domains such as journalism, traffic monitoring, video surveillance, and public safety. In recent years, annotated datasets with extremely dense crowd images have been collected~\cite{zhang2016single,idrees2018composition}, and a number of techniques have been proposed for estimating crowd counts on these challenging datasets. Most recent Crowd Counting approaches~\cite{zhang2016single,sindagi2017generating,sindagi2017cnn,sam2017switching,ranjan2018iterative} estimate a crowd density map, and sum across all the pixels in the crowd density map to obtain the crowd count. A typical crowd counting approach, shown in ~\ref{fig:DensityCNN}, uses a fully convolutional network to map the crowd image into its corresponding crowd density map.

However, the existing convolutional neural network architectures have some limitations for estimating a crowd density map. The convolution operations performed by a neuron in a convolution layer are local, and cannot capture long range dependencies outside the receptive field of the neuron. 
Most neural networks designed for classification tasks such as ~\cite{Krizhevsky-et-al-NIPS12,simonyan2014very, He-et-al-ICCV15} have a few fully connected layers at the end, and each neuron in a fully connected layer has access to the convolutional feature map from the entire image, and hence, can effectively extract non-local information from the convolutional feature maps. Such non-local information is important for making global decisions pertaining to the image, such as predicting the image label. However, the fully convolutional networks typically used for Crowd Counting do not have any fully connected layer to extract non-local features. One possible strategy to extract non-local information in such fully convolutional networks would be to make the networks very deep, so that the convolution filters in the higher layers have a large receptive field. However, training such very deep networks could pose optimization difficulties~\cite{wang2018non}. One can also use fully connected layers, but fully connected layers do not preserve locality, while locality is very important for crowd density estimation at every location.

Our work in this paper is inspired by the Non-local Neural Networks~\cite{vaswani2017attention,wang2018non} that use an attention strategy for capturing long range dependencies without resorting to using very deep convolution layers or fully connected layers.
Using the self-attention mechanism for extracting non-local information,  
we propose an end to end approach which combines local and non-local information for estimating crowd density maps. Note that a few Crowd Counting approaches~\cite{sindagi2017generating} do make use of the non-local information for Crowd Counting. However, they present a multi-stage approach which does not use self-attention, and is very different from our proposed approach. \citet{sindagi2017generating} trained a classifier to classify image patches into different density levels, and used the classifier for extracting non-local information. However, their approach is multi-stage and also requires one to set the different density levels which introduces multiple hyper-parameters into the pipeline.

In this paper, we present an efficient strategy for incorporating non-local information to estimate better quality crowd density maps. Inspired by the Transformer~\cite{vaswani2017attention} and Non-local Neural Network~\cite{wang2018non}, we present Crowd Transformer Network which maps a crowd image to its corresponding density map while taking into account the local as well as non-local information. 

Our key contributions are as follows:
\begin{enumerate}
\item We present the first end-to-end approach for crowd counting which combines local and non-local information for crowd density estimation. 

\item Our proposed approach achieves state-of-art performance on three challenging crowd counting datasets and reduces the mean absolute error by 20 \% and 15 \% on UCF QNRF~\cite{idrees2018composition} and UCF CC~\cite{idrees2013multi} datasets.

\item We propose a novel self-attention mechanism, Contextual Multi-Head Attention, which leads to improvement in the quality of the crowd density maps in comparison to the existing self-attention mechanism~\cite{vaswani2017attention}.

\end{enumerate}

%% file: Related.tex
\section{Related Work}
In this section, we first describe different types of crowd counting approaches, and then discuss the recent CNN based density estimation approaches in more detail. We also describe some of the relevant self-attention mechanisms.

\paragraph{\textbf{{Detection and Regression Based Crowd Counting Approaches.}}}
Some of the earliest crowd counting approaches were detection based~\cite{li2008estimating,lin2001estimation} where the approaches
used classifier based detectors on top of hand crafted feature representation. \citet{lin2001estimation} trained SVM classifier based detectors on top of Haar features for detecting  head. 
\citet{li2008estimating} proposed to first segment the image into crowd and non-crowd regions, and then use a head detector to identify individuals in the crowd region. 
However, these detection-based approaches are severely affected by occlusion, and perform poorly in case of dense crowd images. Some  approaches~\cite{chan2009bayesian,chen2012feature} proposed to circumvent the harder detection problem, and obtain the crowd count by learning a regression function to map image patches to the count in that patch. These regression based approaches also relied on hand crafted features.

\paragraph{\textbf{{Density Estimation Based Crowd Counting Approaches.}}}
\citet{lempitsky2010learning} proposed a density estimating based approach for crowd counting where they learned a linear mapping between crowd patches and the crowd density map. 
\citet{pham2015count} extended the density estimation approach by using a random decision forest for learning the mapping. However, these approaches also use hand crafted features, and do not perform well on the more recent crowd counting datasets~\cite{zhang2016single}. 
More recent approaches of the deep learning era~\cite{zhang2016single,sam2017switching,onoro2016towards,sindagi2017generating,ranjan2018iterative,idrees2018composition,cao2018scale,li2018csrnet} extended the earlier density estimation approaches, and used CNNs for learning the mapping between a crowd image and the corresponding density map. \citet{zhang2016single} pointed out the large variation in crowd densities across different crowd images, and proposed a multi-column architecture where the different columns had filters of varying sizes. The branch with larger kernels could help in extracting features relevant for low density crowd while branch with finer kernel could focus on high density crowd regions. The features from the parallel branches were combined and passed through a $1 \times 1$ convolution layer to predict the crowd density map. One shortcoming of the multi-column approach was that each image, irrespective of its underlying density, was processed by all the columns. \citet{sam2017switching} proposed to  train separate CNN-based density regressors for different density types. They also trained a switch classifier which routed an image patch to the appropriate regressor. \citet{sindagi2017generating} presented an approach which incorporated non-local information for crowd density estimation. They trained an image/patch level classifier for classifying the image/patch into five different density categories. The classifier prediction was used to create context maps which was used along with convolutional features for estimating the crowd density map. \citet{ranjan2018iterative} presented an iterative strategy for crowd density estimation, where a high resolution density map was predicted in two stages. In the first stage, a low resolution density map was estimated, which was used to guide the second stage high resolution prediction. The authors also present a multi-stage generalization of their two stage approach. \citet{lu2018class} presented a class agnostic counting approach, where counting was posed as a matching problem. The authors showed their approach worked not just for human crowd counting, but also for counting objects such as cells and cars.

\paragraph{\textbf{{Self-attention.}}}
Self-attention is an attention mechanism which transforms an input sequence of vectors into another sequence of vectors, where each vector in the output sequence is obtained by relating the corresponding vector in the input sequence with the entire input sequence~\cite{vaswani2017attention}. For handling sequence-to-sequence tasks, \citet{vaswani2017attention}  presented the Transformer architecture which was based entirely on parametric attention modules, and not the usual recurrent neural network architecture. Transformer architecture was initially proposed for handling sequence-to-sequence tasks in the Natural Language Processing domain. \citet{wang2018non} proposed Non-local Neural Networks, which also used self-attention, similar to the Transformer architecture, to extract non-local feature representations. The authors showed that these non-local neural features helped in capturing long range dependencies between features at different spatial/temporal locations which led to improved results for various static image and video classification tasks. The authors also showed how the non-local operations related to the classic non-local means approach~\cite{buades2005non}.

%% file: Proposed.tex
\section{Proposed Approach}

\subsection{Background}\label{sec:Background}
Transformer architecture~\cite{vaswani2017attention} was proposed as an alternative to Recurrent Neural Networks for solving various sequence to sequence tasks in the NLP domain. The architecture consists of an encoder and a decoder where the encoder maps the input sequence into an intermediate representation, which in turn is mapped by the decoder into the output sequence. The transformer uses three types of attention layers: \textit{encoder self-attention, encoder-decoder attention}, and \textit{decoder self-attention}. For the proposed crowd counting approach in this paper, only the first one is relevant which we describe briefly next. Henceforth we will use self-attention  to refer to the self-attention of the encoder. 

\myheading{Encoder Self-Attention:} Given a query sequence along with a key-and-value sequence, the self-attention layer outputs a sequence where the $i$-th element in the output sequence is obtained as a weighted average of the value sequence, and the weights are decided based on the similarity between the $i$-th query element and key sequence. Let $X \in R^{n\times d}$ be a matrix representation of a sequence consisting of $n$ vectors of $d$ dimensions.
The self-attention layer first transforms $X$ into query ($X_Q$), key ($X_K$) and value ($X_V$) matrices by multiplying $X$ with matrices $W_Q$, $W_K$, and $W_V$:
\begin{align}\label{eqn:eqn1}
 &X_Q = X W_Q, \\
 &X_K = X W_K, \\
 &X_V = X W_V.
\end{align}
The output sequence $Z$ is computed efficiently by doing two matrix multiplications:
\begin{equation}\label{eqn:eqn2}
 Z = softmax(X_Q X_K^T)X_V   
\end{equation}
The encoder consists of multiple self-attention layers arranged in a sequential order so that the output of one self-attention layer is fed as input to the next self-attention layer. 

\myheading{Multi-Head Attention:} Instead of a using a single attention head as described above, \citet{vaswani2017attention} suggest to use~$h$ parallel heads, each with its own set of projection matrices. Each  projection matrix is of size $d \times d_k$ and $hd_k = d$. The outputs from the $h$ heads are concatenated, and the resulting matrix is transformed by multiplying with another matrix of size $d \times d$. 
\begin{figure*}[!t]
\centering
\includegraphics[width=\linewidth]{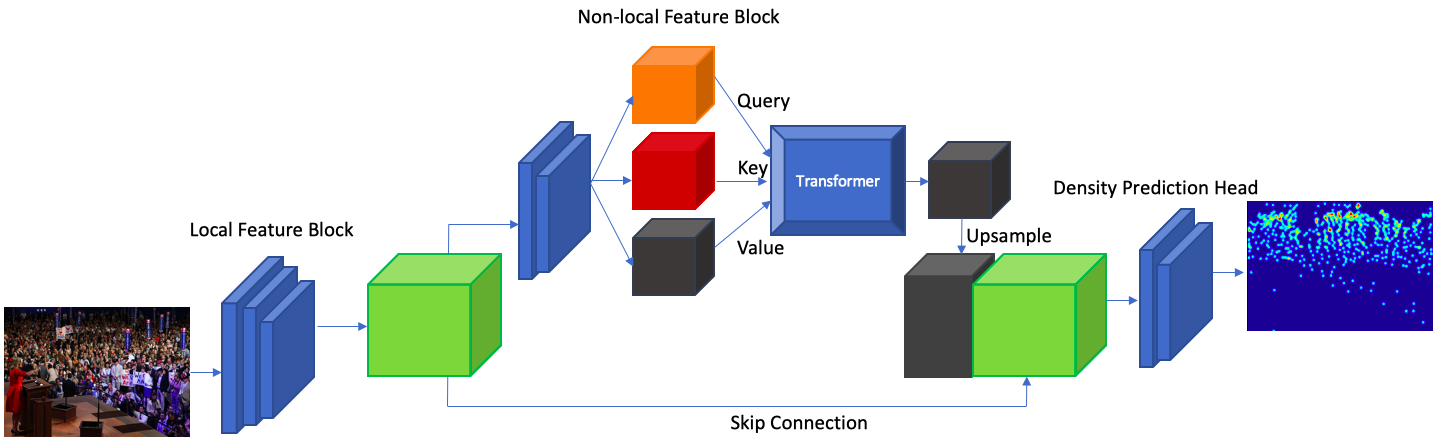}
	\caption{
     \textbf{CTN architecture} combines local and non-local features for crowd density estimation. The local features are computed by the convolution layers in the local feature block. The resulting feature map(shown in green) is passed to the non-local feature block. Non-local feature block uses a novel self-attention mechanism to compute the non-local features(shown in black). Density Prediction Head combines the local and non-local features, and predicts the crowd density map.
     \label{fig:CTN}}	
\end{figure*}

\subsection{Crowd Transformer Network}
We present the block diagram of Crowd Transformer Network (CTN) in Figure~\ref{fig:CTN}.
CTN uses both local and non-local features for estimating the crowd density map. Next, we will describe the Local Feature Block that is used for extracting local features. The Non-local Feature Block, which uses self-attention for extracting non-local features, will be described in Section~\ref{sec:NonLocalFeatureBlock}. In  Section~\ref{sec:DensityPredictionHead}, we will describe the density prediction head, which combines the local and non-local features and predicts a high resolution crowd density map. For non-local feature extraction, we propose a novel attention layer called Contextual Multi-Head Attention, which is described in Section~\ref{sec:ContextualMultiHeadAttention}. 

\subsubsection{Local Feature Block}\label{sec:LocalFeatureBlock}
Given an input image $X$ of size $H \times W$, we pass it through the local feature block to obtain the convolutional feature maps. The local feature block consists of five convolution layers with kernels of size $3 \times 3$, and the number of filters in the convolution  layers are 64, 64, 128, 128 and 256. We use the VGG-16 network~\cite{simonyan2014very} trained on the Imagenet dataset to initialize the convolution layers in local feature block. The local feature block has two max pooling layers, after the second and fourth convolution layer. The resulting feature map is a tensor of size $\frac{H}{4} \times \frac{W}{4} \times 256$. The feature map is passed to non-local block as well as the density prediction head. 

\subsubsection{Non-local Feature Block}\label{sec:NonLocalFeatureBlock}
The non-local feature block takes as input the feature map from the local feature block, and passes it through 3 convolution layers of kernel size $3 \times 3$ and a max pooling layer, which results in a feature map of size $\frac{H}{8} \times \frac{W}{8} \times 512$. We reduce the depth of the feature map by passing it through a $1 \times 1$ convolution layer, which yield a feature map of size $\frac{H}{8} \times \frac{W}{8} \times 240$. The resulting feature map is flattened into a matrix of size $M \times 240$, where $M = \frac{H}{8} \times \frac{W}{8}$. Each row in this matrix corresponds to some location in the convolution feature map. The flattened matrix is passed through three self-attention layers. The output from final transformer layer is reshaped back into a tensor of size $\frac{H}{8} \times \frac{W}{8} \times 240$. 

The multi-head attention layer, while transforming the input sequence into the query, key and value sequences(as described in Section~\ref{sec:Background}), does not utilize any context information. We present a novel multi-head attention mechanism, called Contextual Multi-Head Attention, that uses context information while transforming an input sequence. We will describe the Contextual Multi-Head Attention layer in Section~\ref{sec:ContextualMultiHeadAttention}

\subsubsection{Density Prediction Head}\label{sec:DensityPredictionHead}
Both local and non-local features are important for estimating an accurate crowd density map. Hence, the Density Prediction Head uses a skip connection to obtain the convolutional features from the local feature block, and combines it with the features from the non-local feature block. The non-local features are upsampled to the same spatial size as local features, which results in a tensor of size $\frac{H}{4} \times \frac{W}{4} \times 240$. The local and non-local features are concatenated and passed through 4 convolution layers (with 196, 128, 48, and 1 filters), where the last layer is a $1 \times 1$ convolution layer. We add a ReLU non-linearity after the $1 \times 1$ convolution layer to prevent the network from predicting negative density. We use two bilinear interpolation layers, after the second and third convolution layers in the prediction head. Each of the interpolation layer upsamples its input to twice its size. The input to the final $1\times1$ convolution layer is a feature map of size $H \times W \times 48$, which is transformed into a 2D map by the last convolution layer.
\subsection{Contextual Multi-Head Attention}\label{sec:ContextualMultiHeadAttention}
As described in Section~\ref{sec:Background}, the self-attention layer in the Transformer architecture~\cite{vaswani2017attention} first transforms the input sequence into query, key, and value sequences by multiplying each vector in the input sequence with a weight matrix before applying the attention mechanism. Let $x_i$ the $i$-th vector in matrix $X$. $X$ here is the input to the self-attention layer. We also look at the immediate neighborhood of $x_i$, i.e. $x_{i-1}$ and $x_{i+1}$, while transforming $x_i$ into $x_i^Q$,  $x_i^K$,  $x_i^V$:
\begin{eqnarray}\label{eqn:eqn3}
  x_i^Q = concat(x_{i-1},x_i,x_{i+1}) W_Q, \\
  x_i^K = concat(x_{i-1},x_i,x_{i+1}) W_K,  \\
  x_i^V = concat(x_{i-1},x_i,x_{i+1}) W_V,  
\end{eqnarray}
where $concat(\cdot,\cdot,\cdot)$ is a function that output the concatenated vector for the input vectors. The above equation can be implemented efficiently by using a 1D convolution layer:
\begin{eqnarray}\label{eqn:eqn4}
  X_Q = f_{1D-Conv}(X, \theta_Q), \\
  X_K = f_{1D-Conv}(X, \theta_K),  \\
  X_V = f_{1D-Conv}(X, \theta_V),  
\end{eqnarray}
where $f_{1D-Conv}(X,\theta$ refers to the 1D convolution layer with parameter $\theta$ and $X$ is the input sequence. We pad the input sequence with zero vectors to handle the boundary cases. The query, key, and value sequences are transformed into the output sequence: $Z = softmax(X_QX_K^T)X_V$.
Similar to the Transformer architecture~\cite{vaswani2017attention}, we use $h$ parallel attention heads and concatenate the $h$ resulting sequences into a single sequence, which is transformed by another 1D convolution layer into the output sequence. For our experiments, we use $3$ such contextual multi-head attention layers for extracting the non-local features. 

Instead of depending on just the immediate neighbors of $x_i$, we can use a larger context. Below we describe the equation to transform $x_i$ using $2m$ context vectors:
\begin{eqnarray}\label{eqn:eqnA1}
  x_i^Q = concat(x_{i-m},\ldots,x_{i-1},x_i,x_{i+1},\ldots,x_{i+m}) W_Q, \\
  x_i^K = concat(x_{i-m},\ldots,x_{i-1},x_i,x_{i+1},\ldots,x_{i+m}) W_K,  \\
  x_i^V = concat(x_{i-m},\ldots,x_{i-1},x_i,x_{i+1},\ldots,x_{i+m}) W_V,  
\end{eqnarray}
where $concat(\cdot,\cdot,\cdot)$ is a function that output the concatenated vector for the input vectors. Note that the size of the transformation matrices depend upon the context, the input dimensionality and the dimensionality of the resulting vector. The above equation can be implemented efficiently by using a 1D convolution layer:
\begin{eqnarray}\label{eqn:eqnA2}
  X_Q = f_{1D-Conv}(X, \theta_Q), \\
  X_K = f_{1D-Conv}(X, \theta_K),  \\
  X_V = f_{1D-Conv}(X, \theta_V), 
\end{eqnarray}
where $f_{1D-Conv}(X,\theta)$ refers to the 1D convolution layer with parameters $\theta$ and $X$ is the input sequence. 

For our experiments, we set the number of parallel attention heads $h$ to 12.

%% file: Results.tex
\section{Experiments}
We conduct experiments on three challenging datasets: UCF-QNRF~\cite{idrees2018composition}, Shanghaitech~\cite{zhang2016single}, and UCF Crowd Counting~\cite{idrees2013multi}.

\subsection{Evaluation Metrics}

Following the previous Crowd Counting works, we compare the proposed approach to the existing approaches in terms of Mean Absolute Error (MAE) and Root Mean Squared Error (RMSE). If the predicted count for an image is $\hat{y}$ and
the ground truth count is $y$, the MAE and RMSE can be computed as: 
\begin{eqnarray}
&MAE = \frac{1}{n}\sum_{i=1}^{n} \lvert y_i - \hat{y_i} \rvert, \\
&RMSE = \sqrt[]{\frac{1}{n}\sum_{i=1}^{n} (y_i - \hat{y_i})^2},
\end{eqnarray}
where the summation is done over the test set.

\subsection{Results on the UCF-QNRF dataset}
\begin{table}[!t]
\centering

\begin{tabular}{lrr}
\toprule
Method     & MAE   & RMSE   \\
\midrule
Idrees \textit{et al.}~\cite{idrees2013multi} & 315 & 508  \\
MCNN~\cite{zhang2016single} & 277 & 426 \\
Encoder-Decoder~\cite{badrinarayanan2017segnet} & 270 & 478 \\
CMTL~\cite{sindagi2017cnn}& 252 & 514 \\
Switch CNN ~\cite{Sam-etal-CVPR17} & 228 & 445 \\
Resnet101~\cite{he2016deep} & 190 & 277 \\
Densenet201~\cite{huang2017densely} & 163 & 226 \\
Composition Loss-CNN~\cite{idrees2018composition} & 132 & 191 \\
CTN (Proposed)  &\underline{\textbf{102.6}} & \underline{\textbf{177.7}} \\
\bottomrule
\end{tabular}
\vskip 0.1in
\caption{{\bf Performance of various methods on the UCF-QNRF dataset}. We compare our proposed approach with the previous approaches in terms of MAE and RMSE metrics. Training is done on random squared crops of size 384. The proposed approach outperforms all other approaches by a large margin in terms of both the MAE and the RMSE metrics. Note that we run the experiment twice, and report the average MAE and RMSE values for CTN. Numbers corresponding to CTN are highlighted, and the numbers from the best approach are underlined.  \label{tab:tableucf}}
\end{table}

UCF-QNRF~\cite{idrees2018composition} is the largest annotated crowd counting dataset with 1535 crowd images and 1.2 million annotations. Images in the dataset were collected by doing image search on Google and Flickr search engines and from Hajj footage. The dataset is divided into 12,01 training and 334 test images. Crowd count across the images varies between 49 and 12,865. Each person in the dataset is annotated with a single dot annotation, and this dot annotation map is convolved with a Gaussian to obtain the target density map for training the crowd density estimation network.

We first re-scale those images in the dataset where the larger dimension is greater than 1920. We preserve the original aspect ratio while scaling the images. For training, we take 100 random crops of size $384\times384$ from each of the training images. We use a batchsize of 3 and train our model for 10 epochs. We use a learning rate of $10^{-4}$ and  Adam optimizer for training our model. We normalize the training and test images using the mean and variance values computed on the ImageNet dataset. We initialize the five convolutional layers in the local feature block  by using the  first 5 layers from a Vgg Net~\cite{simonyan2014very} trained on the ImageNet dataset. We initialize the first convolution layers in the non-local block (except the $1\times1$ convolution layer and the 1D convolution layer in the Contextual Multi-Head Attention layer) using the pretrained layers from Vgg16. The remaining learnable parameters are initialized randomly using a Gaussian distribution with zero mean and standard deviation of $0.001$. Since we perform the evaluation on a single GPU, we run into memory issue for larger test images. For such images, we divide the image into non-overlapping crops, obtain the density map for each crop,  and sum the count across all the crops to obtain the overall count.
In Table~\ref{tab:tableucf}, 
we compare CTN with the previous state-of-art methods in terms of MAE and RMSE metrics. CTN outperforms all previous methods by a large margin in terms of both the MAE and the RMSE metrics.


\subsection{Ablation Study on UCF QNRF}
In Table~\ref{tab:ablation}, we show the results for ablation study conducted on the UCF-QNRF dataset~\cite{idrees2018composition}. We analyze the importance of the key components of the proposed CTN architecture: 1) local features, 2) non-local features, and 3) Contextual Multi-Head Attention. We observe that removing either the local or the non-local feature results in a drastic drop in performance. This observation shows that both the local and the non-local features are needed for predicting accurate density map. We also observe that using Multi-Head Attention~\cite{vaswani2017attention}, instead of using the proposed Contextual Multi-Head Attention mechanism, leads to worse results. 
\begin{table}[!t]
\centering
\begin{tabular}{lrr}
\toprule
Method     & MAE   & RMSE   \\
\midrule
Local features &120.2 &218.4 \\
Non-local features &123.5 &206.7 \\
CTN, but with Multi-Head
Attention &108.3 &190.8 \\
CTN (proposed) &102.6 & 177.7\\
\bottomrule
\end{tabular}
\vskip 0.1in
\caption{{\bf Ablation study on the UCF-QNRF dataset~\cite{idrees2018composition}}. The \textit{Local features} approach does not use the non-local feature block and uses only the local features for estimating the crowd count. The \textit{Non-local features} approach removes the skip connection from the CTN architecture, i.e., the prediction head uses only the non-local features. The third method is \textit{CTN}, but with Multi-Head Attention mechanism proposed by ~\citet{vaswani2017attention}; \textit{CTN} refers to the proposed architecture.
\textit{Local features} and \textit{Non-local features} approaches perform poorly in comparison to other two approaches. This suggests that both local and non-local features are important for accurate crowd density estimation. CTN outperforms CTN + Multi-Head Attention which shows the benefits of using the contextual attention mechanism.
\label{tab:ablation}}
\end{table}

\subsection{Effect of Varying Context on UCF-QNRF dataset}\label{sec:EffectofVaryingContextonUCF-QNRFdataset}
To analyze the effects of varying the context information, we conduct experiments on the UCF-QNRF dataset~\cite{idrees2018composition}. We vary the context,i.e. the number of adjacent vectors in Equations \textit{1-3} in the supplementary submission, and analyze the impact on the quality of the estimated crowd density map. In Table~\ref{tab:ablationContext}, we present the results on UCF-QNRF~\cite{idrees2018composition}. The context $0$ refers to the original Multi-Head Attention mechanism proposed by Vaswani \etal~\cite{vaswani2017attention}. Context $2$ are Contextual Multi-Head Attention results presented in the main paper. From the table, we can see that using more context helps. We see that using a context of $6$ significantly outperforms the attention mechanism presented in the Transformer architecture~\cite{vaswani2017attention}. Furthermore using a context of $6$ also leads to improvements over using a context of $2$ from the main paper. We do not experiment with context larger than 10 since the performance plateaus for context larger than 6.
\begin{table}[t]
\centering
\begin{tabular}{crr}
\toprule
 Context & MAE   & RMSE   \\
\midrule
 0& 108.3 & 190.8 \\
 2 & 105.7 & 184.5\\
 4 & 104.0 & 183.0 \\
 6 & \textbf{102.6} & 177.7 \\
 10 & 103.0  & \textbf{176.0} \\
\bottomrule
\end{tabular}
\vskip 0.1in
\caption{{\bf Effect of varying Context in the CTN arhictecture on UCF-QNRF dataset~\cite{idrees2018composition}}:In this table, we vary the context information used in our proposed Contextual Multi-Head Attention layer. Context refers to the 2m context vectors in Equations \textit{1 - 3}. Context of 0 corresponds to the attention mechanism proposed by Vaswani \etal~\cite{vaswani2017attention} for which only the vector at current location is taken into consideration while transforming the sequence. Context of 2 are the CTN results presented in the main paper. In terms of MAE metric, performance keeps improving as we increase the context from 0  to 6.
\label{tab:ablationContext}}
\end{table}


\subsection{Results on UCF-CC dataset}
\begin{table}[!t]
\centering

\begin{tabular}{lrr}
\toprule
Method     & MAE   & RMSE   \\
\midrule
Lempitsky \& Zisserman~\cite{lempitsky2010learning}  & 493.4 & 487.1 \\
Idrees et. al~\cite{idrees2013multi}     &  419.5     & 487.1      \\
Crowd CNN ~\cite{zhang2015cross}     &    467.0   & 498.5      \\
Crowdnet~\cite{boominathan2016crowdnet}   &     452.5  & -      \\
MCNN~\cite{zhang2016single}      &   377.6    &  509.1     \\
Hydra2s ~\cite{onoro2016towards}  &    333.7   &  425.6    \\
Switch CNN~\cite{sam2017switching} &     318.1  &  439.2     \\
CP-CNN~\cite{sindagi2017generating} &295.8 & \underline{320.9} \\
IG-CNN~\cite{babu2018divide}  & 291.4 & 349.4 \\
ic-CNN ~\cite{ranjan2018iterative} & 260.9 & 365.5 \\
SANet ~\cite{cao2018scale} & 258.4 & 334.9 \\
CSR Net ~\cite{li2018csrnet} & 266.1 & 397.5 \\
CTN(Proposed) & \underline{\textbf{219.3}} & \textbf{331.0}\\
\bottomrule
\end{tabular}
\vskip 0.1in
\caption{{\bf Performance of various methods on the UCF Crowd Counting dataset.} We compare our proposed approach with the previous approaches in terms of MAE and RMSE metrics. We perform five fold cross validation and report the average across the five runs. Training is done on random crops of size $\frac{H}{3} \times \frac{W}{3}$. The proposed CTN  outperforms all other approaches by a large margin in terms of MAE metric. In terms of RMSE metric, our proposed approach outperforms all the methods except CP-CNN~\cite{sindagi2017generating}. Numbers corresponding to CTN are highlighted, and the numbers from the best approach are underlined.  \label{tab:tableucf2}}
\end{table}

\setlength{\tabcolsep}{14pt}
\begin{table*}[!hbt]
\begin{center}	

\begin{tabular}{lrrrr}
\toprule
         & \multicolumn{2}{c}{Part A} & \multicolumn{2}{c}{Part B} \\
         \cmidrule(lr){2-3} \cmidrule(lr){4-5} 
               & MAE          & RMSE         & MAE          & RMSE         \\
\midrule
Crowd CNN~\cite{zhang2015cross}     &        181.8      &  277.7           &   32.0           &  49.8           \\


MCNN~\cite{zhang2016single}           &         110.2     &    173.2         &  26.4            &  41.3           \\
Switching CNN~\cite{sam2017switching}  &   90.4           &  135.0           &     21.6         &     33.4        \\
CP-CNN~\cite{sindagi2017generating}  & 73.6 & 106.4 &20.1 & 30.1  \\
IG-CNN~\cite{babu2018divide} & 72.5 & 118.2 & 13.6 & 21.1 \\
ic-CNN~\cite{ranjan2018iterative}   & 68.5 &116.2 & 10.7& 16.0\\
SANet ~\cite{cao2018scale} &  67.0 & \underline{104.5} & \underline{8.4} & \underline{13.6} \\
CSR Net ~\cite{li2018csrnet} & 68.2 & 115.0 & 10.6 & 16.0 \\
CTN(Proposed) & \underline{\textbf{64.3}} & \textbf{107.0} & \underline{\textbf{8.6}} & \textbf{14.6} \\
\bottomrule
\end{tabular}
\end{center}
\vskip 0.1in
\caption{{\bf Count errors of different methods on the Shanghaitech dataset.}  This dataset has two parts: A and B. Images in Part A are collected from the web, while those in Part B are collected on the streets of Shanghai. Part A images are typically more dense than Part B images. The average error on Part B is less compared to that in Part A . We compare our proposed approach with the previous approaches in terms of MAE and RMSE metrics. Training is done on random crops of size $\frac{H}{3} \times \frac{W}{3}$. Numbers corresponding to CTN are highlighted, and the numbers from the best approach are underlined.  \label{table_shanghaitech}
}
\end{table*}

The UCF Crowd Counting dataset~\cite{idrees2013multi} contains $50$ crowd images with widely varying crowd count. The images are collected via web search. Each person in the dataset is annotated with a single dot annotation, and this dot annotation map is convolved with a Gaussian to obtain the target density map for training. The maximum, minimum and average counts are $94$, $4545$ and $1280$ respectively. Since the dataset is small in comparison to the UCF-QNRF dataset, we use the CTN network trained on UCF-QNRF to initialize the model. We train the network using random crops of sizes $\frac{H}{3}\times\frac{W}{3}$.
Following the evaluation protocol in earlier works~\cite{zhang2016single},  we perform five-fold cross validation and report the average MAE and RMSE results in Table ~\ref{tab:tableucf2}. The proposed approach outperforms all other approaches by a large margin in terms of MAE metric. In terms of RMSE metric, CTN outperforms all the methods except CP-CNN~\cite{sindagi2017generating}.


\subsection{Results on Shanghaitech}
The Shanghaitech crowd counting dataset contains of two parts. Part-A consists of 482 images collected from the web, and Part B consists of 716 images collected on the streets of Shanghai. 
Images for Part A are typically denser than Part B images.
 Each person in the dataset is annotated with a single dot annotation, and this dot annotation map is convolved with a Gaussian to obtain the target density map for training. Since the dataset is small in comparison to the UCF-QNRF dataset, we use the CTN network trained on UCF-QNRF for initializing the model. We train the network using random crops of sizes $\frac{H}{3}\times\frac{W}{3}$. 
 
 In Table~\ref{table_shanghaitech}, we compare the proposed approach with the previous approaches in terms of MAE and RMSE metrics. CTN outperforms all the approaches approaches except SANet~\cite{cao2018scale} on both Part-A and Part-B datasets. On Part A data, CTN outperforms SANet in terms of MAE metric.



\subsection{Qualitative Results}
\begin{figure*}
\begin{center}
\begin{tabular}{ccc} 

{\bf Image} & {\bf Ground truth} & {\bf Output}   \\ \\ 
{  
	\includegraphics[scale=0.25]{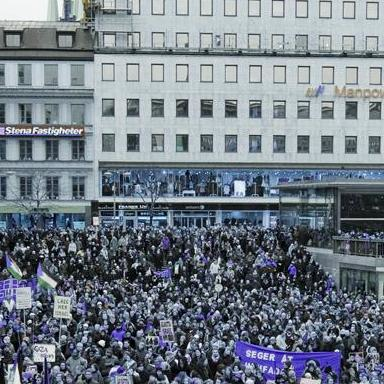}
     } &  {  
	\includegraphics[scale=0.25]{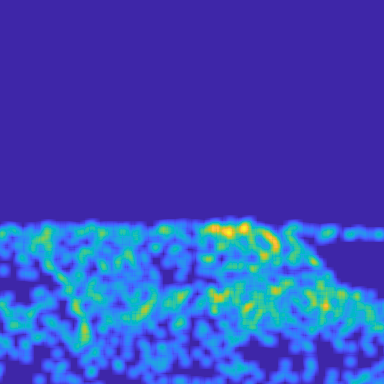}
     }
     &
     {\includegraphics[scale=0.25]{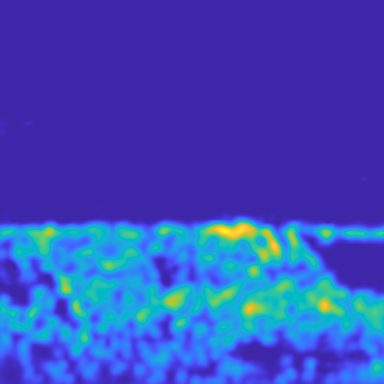}
     } \\
    &645  & 647 \\
   \\ 
{  
	\includegraphics[scale=0.25]{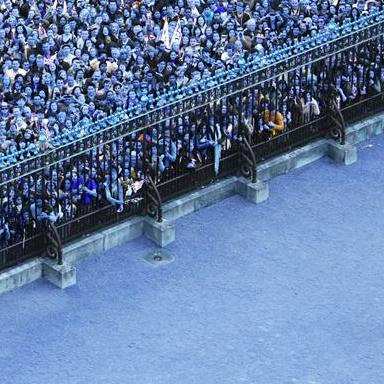}
     } &  {  
	\includegraphics[scale=0.25]{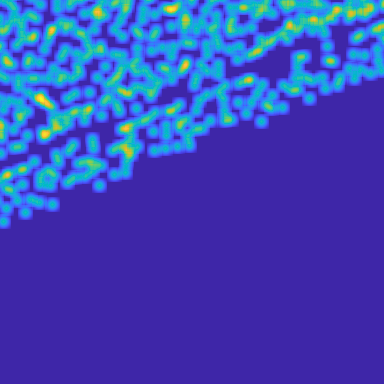}
     }
     &
     {\includegraphics[scale=0.25]{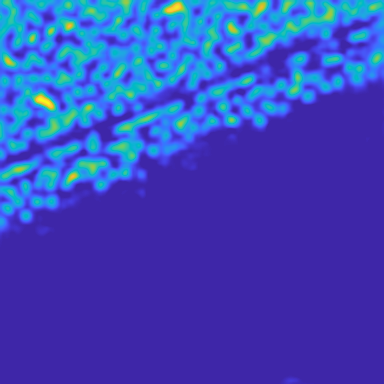}
     } \\
    &349  & 328 \\
   \\ 
{  
	\includegraphics[scale=0.25]{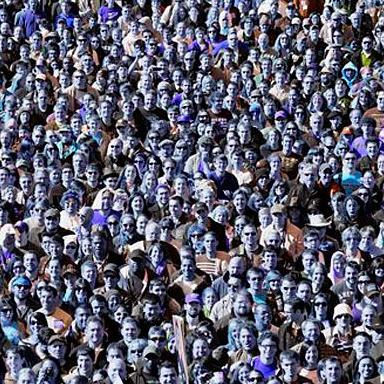}
     } &  {  
	\includegraphics[scale=0.25]{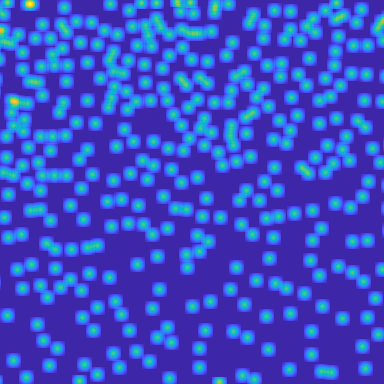}
     }
     &
     {\includegraphics[scale=0.25]{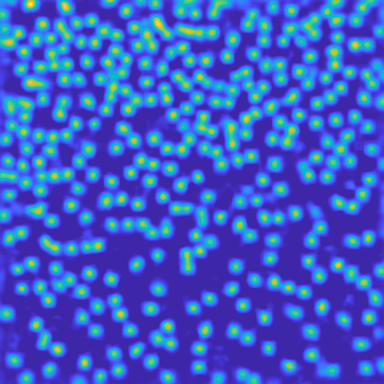}
     } \\
    &347  & 354 \\
   \\ 
{  
	\includegraphics[scale=0.25]{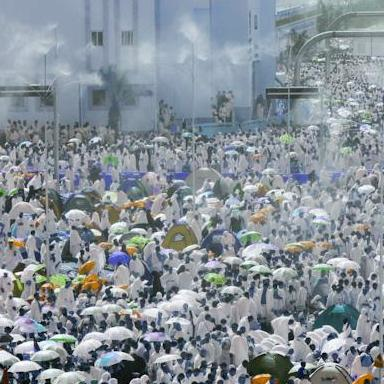}
     } &  {  
	\includegraphics[scale=0.25]{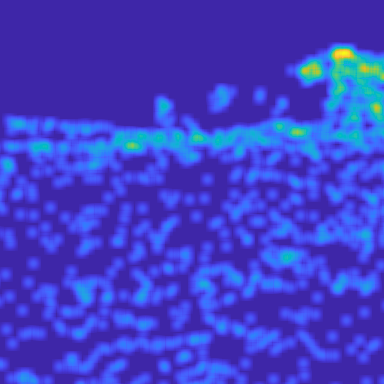}
     }
     &
     {\includegraphics[scale=0.25]{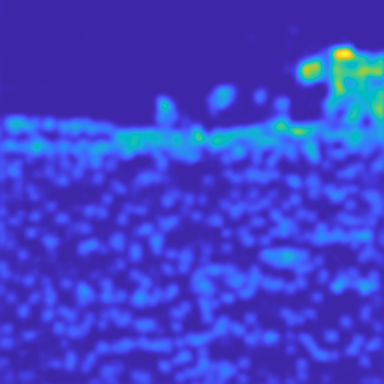}
     } \\
    &841  & 792 \\
   \\ 
{  
	\includegraphics[scale=0.25]{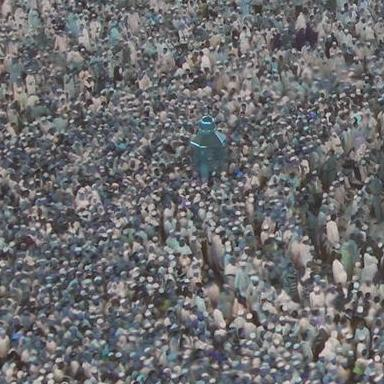}
     } &  {  
	\includegraphics[scale=0.25]{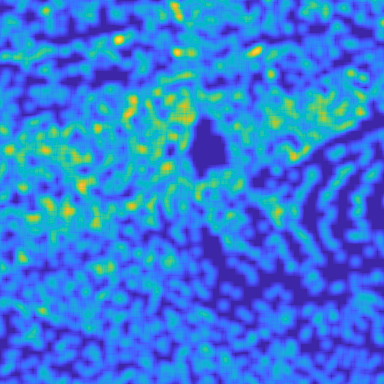}
     }
     &
     {\includegraphics[scale=0.25]{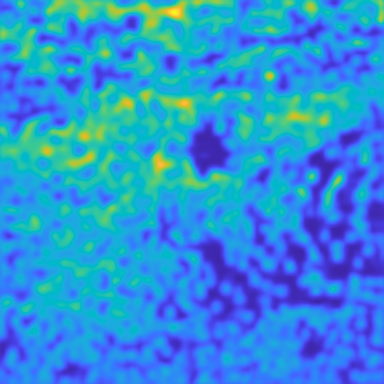}
     } \\
    &1703  & 1586 \\
   \\ 

\end{tabular}
\caption{{\bf Qualitative results, some success and failure cases.} The three columns show the input image, ground truth annotation map, and the prediction. The total counts are shown below each density map. The first three rows are success cases for CTN, while the last two are failure cases. \label{fig:quali}}
\end{center}
\end{figure*}

In Figure~\ref{fig:quali}, we show a few qualitative results obtained using our proposed approach. The three columns show the input image, ground truth annotation map, and the predicted density map. The first three rows are success cases for CTN, while the last two are failure cases. Among the results shown, the error is the largest for the last crowd image which is the most dense image among the ones shown. 
\setlength{\tabcolsep}{20pt}

\subsection{Local Vs Non-local Features}\label{sec:qualitativeB}
\begin{figure*}
\begin{center}
\begin{tabular}{cccc} 

{\bf Image} & {\bf Ground truth} & {\bf CTN(Local + Non-local)} &  {\bf Local}   \\ \\ 

{  
	\includegraphics[scale=0.17]{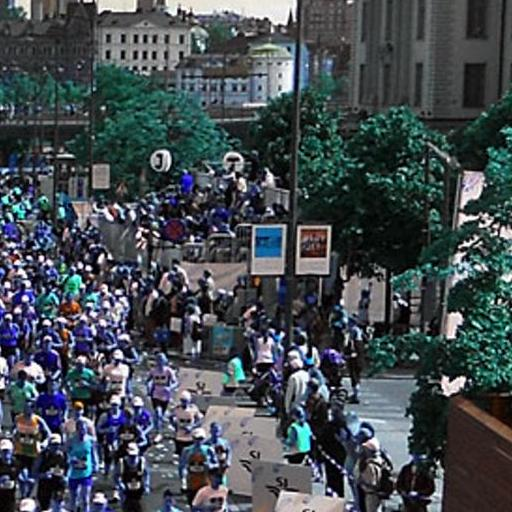}
     } &  {  
	\includegraphics[scale=0.17]{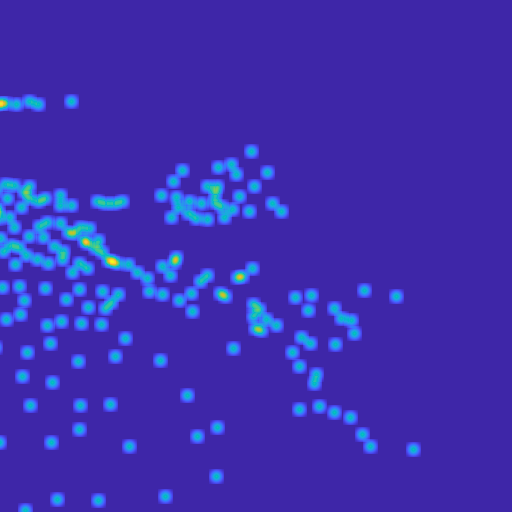}
     }
     &
     {\includegraphics[scale=0.17]{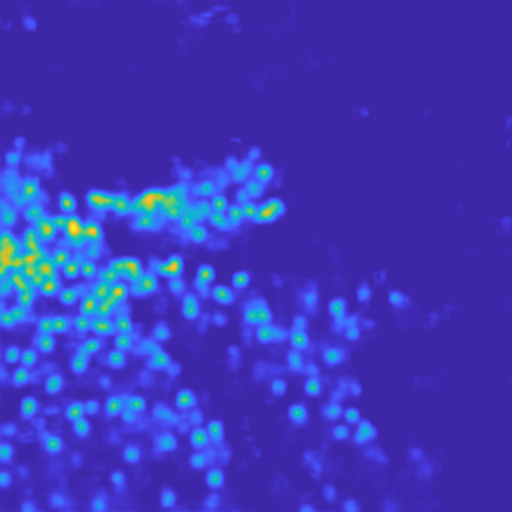}
     } 
          &
     {\includegraphics[scale=0.17]{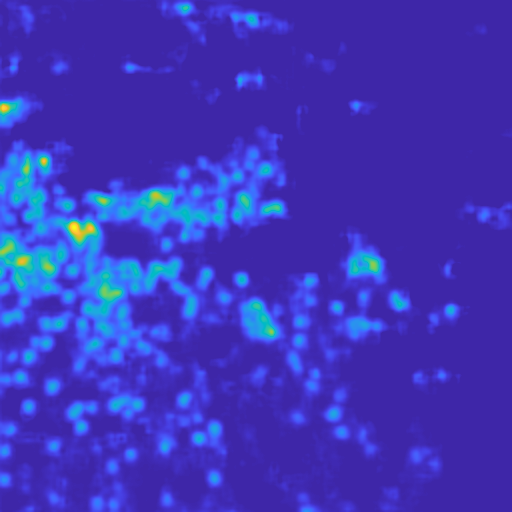}
     }
     \\
    &172 & 207 & 291 \\
   \\ 
   
{  
	\includegraphics[scale=0.17]{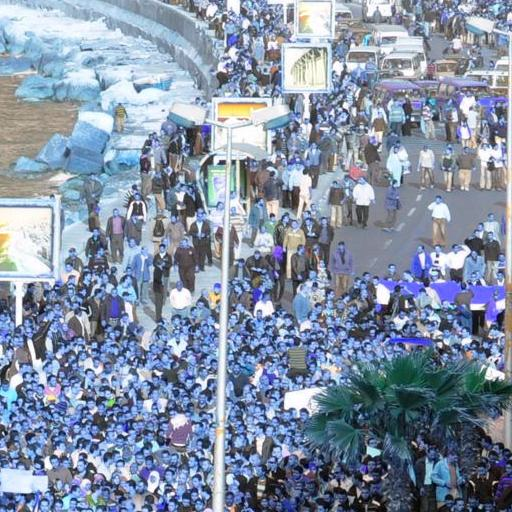}
     } &  {  
	\includegraphics[scale=0.17]{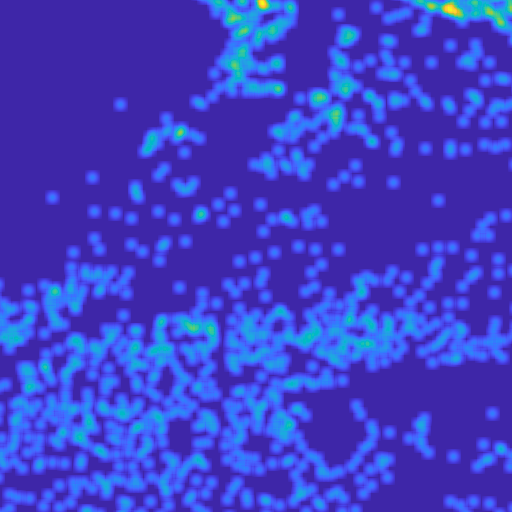}
     }
     &
     {\includegraphics[scale=0.17]{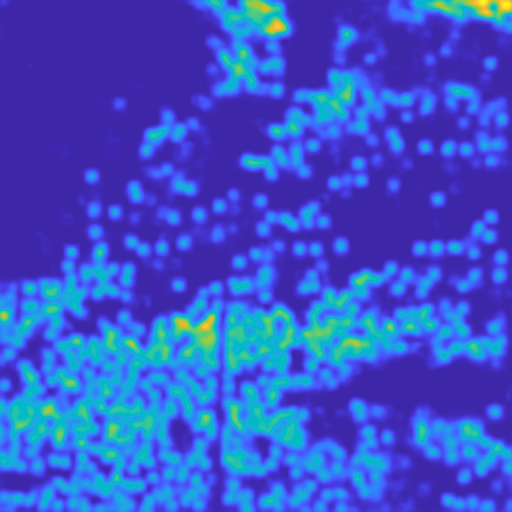}
     } 
          &
     {\includegraphics[scale=0.17]{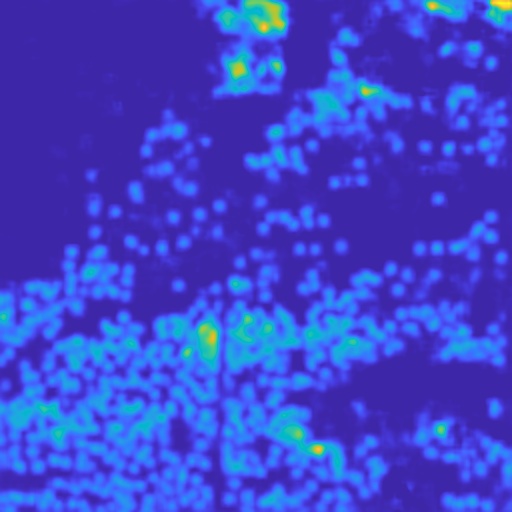}
     }
     \\
    &954 & 921 & 849\\
   \\

   {  
	\includegraphics[scale=0.17]{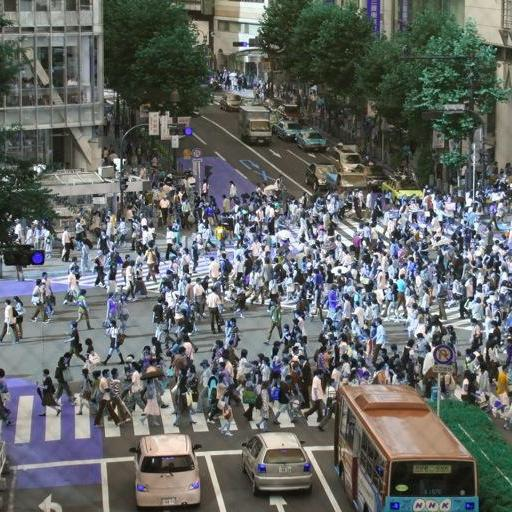}
     } &  {  
	\includegraphics[scale=0.17]{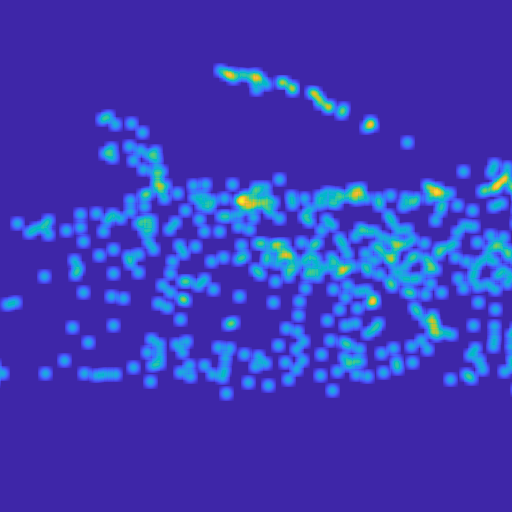}
     }
     &
     {\includegraphics[scale=0.17]{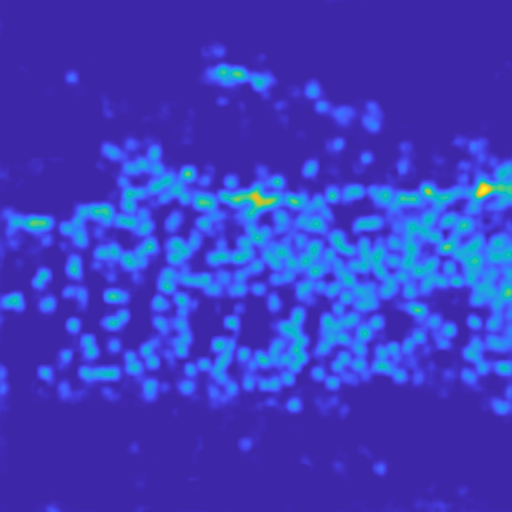}
     } 
          &
     {\includegraphics[scale=0.17]{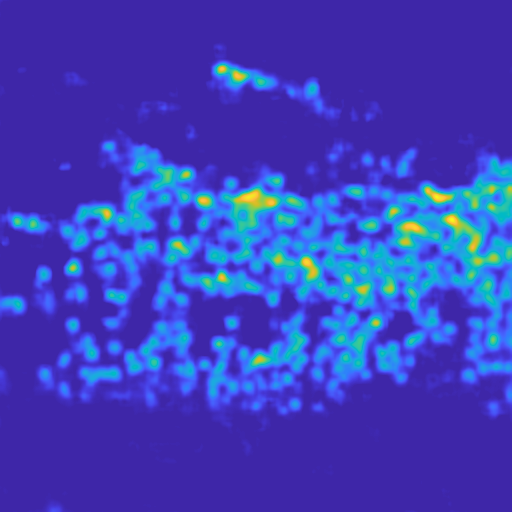}
     }
     \\
    & 438 & 470 & 505 \\
   \\ 
 
{  
	\includegraphics[scale=0.17]{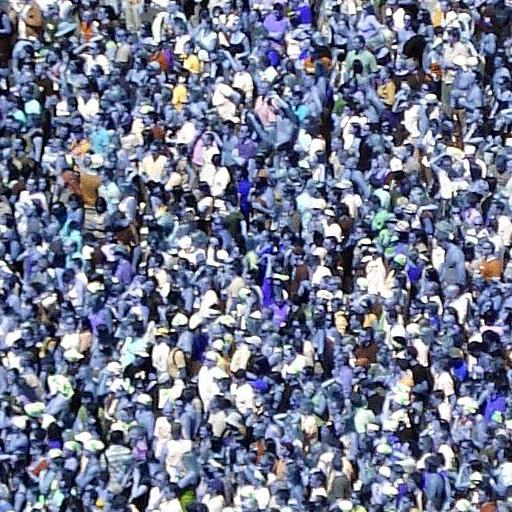}
     } &  {  
	\includegraphics[scale=0.17]{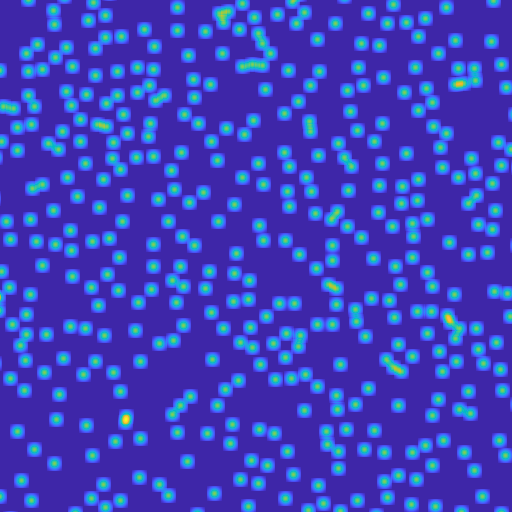}
     }
     &
     {\includegraphics[scale=0.17]{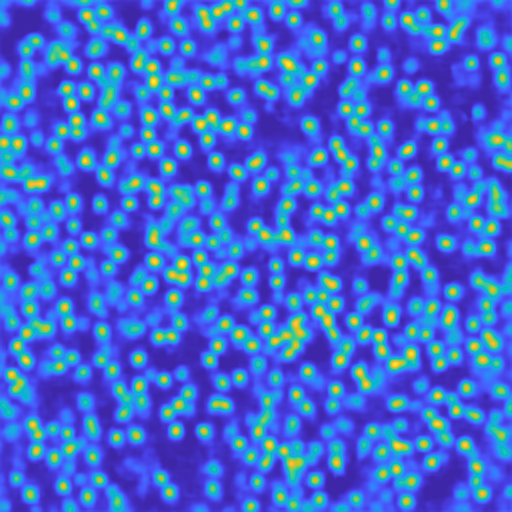}
     } 
          &
     {\includegraphics[scale=0.17]{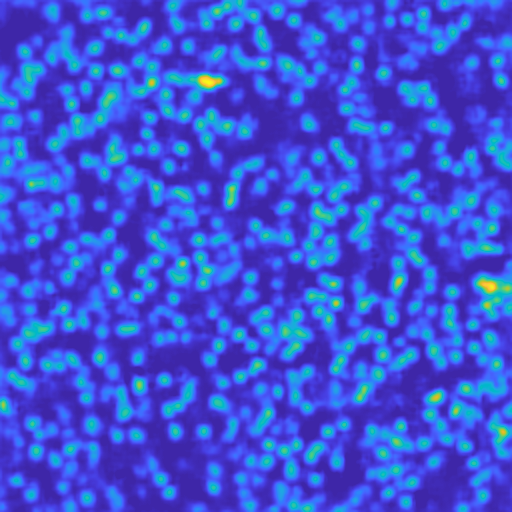}
     }
     \\
    &405 & 548 & 581 \\
   \\   
{  
	\includegraphics[scale=0.17]{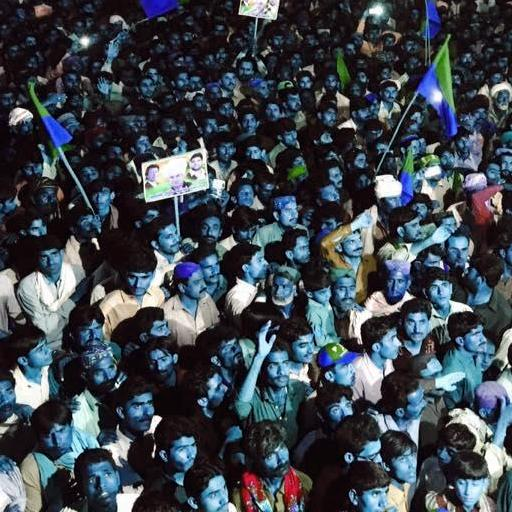}
     } &  {  
	\includegraphics[scale=0.17]{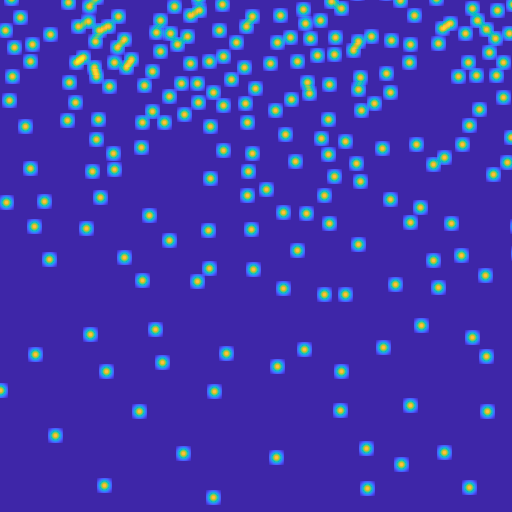}
     }
     &
     {\includegraphics[scale=0.17]{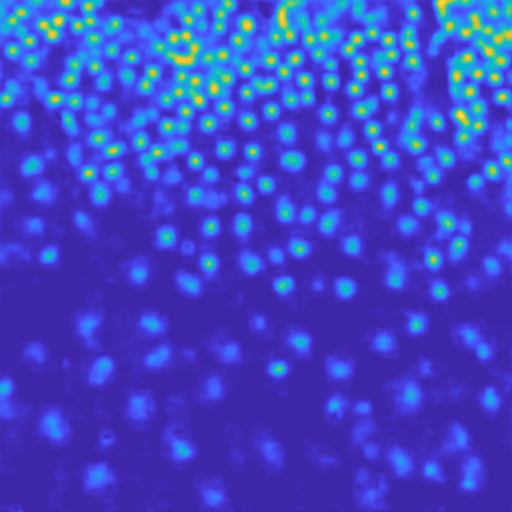}
     } 
          &
     {\includegraphics[scale=0.17]{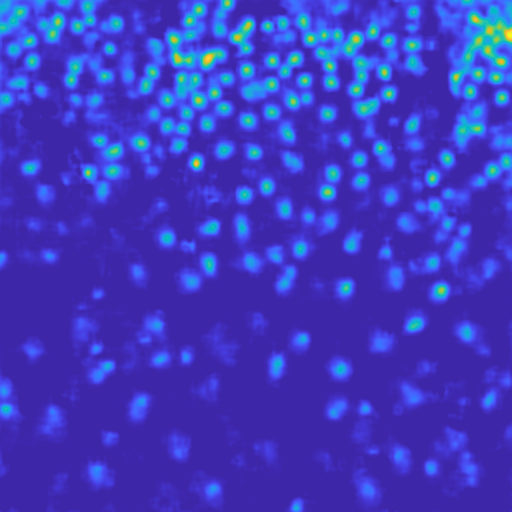}
     }
     \\
    & 208 & 276 & 222 \\
   \\ 
   \\ 
\end{tabular}
\caption{{\bf Qualitative results, some success and failure cases.} The four columns show the input image, ground truth annotation map, the CTN prediction and the Local prediction. The total counts are shown below each density map. The first three rows are success cases for CTN, while the last two are failure cases for CTN. For the last row, the local approach performs better than CTN. \label{fig:qualiCTNvsLocal}}
\end{center}
\end{figure*}

In this section, we do a qualitative analysis of the benefits of combining non-local features with local features for crowd density estimation. We compare the proposed CTN approach, which combines local and non-local features, with a local approach, which uses only the convolutional features for estimating the crowd density map. For the local baseline, we remove the non-local block from the CTN architecture. We show some qualitative results on UCF-QNRF~\cite{idrees2018composition} dataset in Figure~\ref{fig:qualiCTNvsLocal}.

We compare the two approaches and present few success and failure cases. We show the crowd image, corresponding ground truth density map and the predicted crowd density map obtained via both of the approaches. The first three examples are success cases for CTN, while the last two are failure cases. For majority of examples shown in the figure, predictions obtained using the proposed CTN approach are more accurate than those obtained by the local approach. For the first example, the density map from the local approach contains false positives towards the top while the CTN prediction contains significantly less false positives. The second and third examples comprise of dense crowd, where the people in the crowd occupy a small number of pixels. CTN significantly outperforms the local approach in this case. This suggests that the non-local features help in better handling very dense crowd images. The last two examples are failure cases for both of the approaches.

